\documentclass{article}
\usepackage{graphics,graphicx,caption,float,subcaption,booktabs,xcolor,multirow,array,color,ifthen,tabu,colortbl,dblfloatfix,url,relsize,algorithm,algorithmic,amssymb,xspace,nicefrac,microtype,amsmath,amsfonts,bm}
\usepackage[pagebackref=true,breaklinks=true,colorlinks=true,bookmarks=false]{hyperref}
\usepackage[capitalise,noabbrev,nameinlink]{cleveref}
\usepackage[english]{babel}
\usepackage[accepted]{icml2021}
\usepackage{enumitem}
\usepackage{mathtools}

\definecolor{algo_color}{RGB}{157, 36, 92}

\icmltitlerunning{Unsupervised Learning of Visual 3D Keypoints for Control}
\begin{document}
\twocolumn[
\icmltitle{Unsupervised Learning of Visual 3D Keypoints for Control}
\icmlsetsymbol{equal}{*}
\begin{icmlauthorlist}
\icmlauthor{Boyuan Chen}{ucb}
\icmlauthor{Pieter Abbeel}{ucb}
\icmlauthor{Deepak Pathak}{cmu}
\end{icmlauthorlist}
\icmlaffiliation{ucb}{UC Berkelely}
\icmlaffiliation{cmu}{Carnegie Mellon University}
\icmlcorrespondingauthor{Deepak Pathak}{dpathak@cs.cmu.edu}
\icmlkeywords{keypoints, 3D vision, reinforcement learning, control, unsupervised learning, robotics, manipulation, locomotion, deformable}
\vskip 0.3in
]
\printAffiliationsAndNotice{}

\begin{abstract}
\begin{hyphenrules}{nohyphenation}
Learning sensorimotor control policies from high-dimensional images crucially relies on the quality of the underlying visual representations. Prior works show that structured latent space such as visual keypoints often outperforms unstructured representations for robotic control. However, most of these representations, whether structured or unstructured are learned in a 2D space even though the control tasks are usually performed in a 3D environment. In this work, we propose a framework to learn such a 3D geometric structure directly from images in an end-to-end unsupervised manner. The input images are embedded into latent 3D keypoints via a differentiable encoder which is trained to optimize both a multi-view consistency loss and downstream task objective. These discovered 3D keypoints tend to meaningfully capture robot joints as well as object movements in a consistent manner across both time and 3D space. The proposed approach outperforms prior state-of-art methods across a variety of reinforcement learning benchmarks. Code and videos at~\url{https://buoyancy99.github.io/unsup-3d-keypoints/}.
\end{hyphenrules}
\end{abstract}

\section{Introduction}
\label{sec:introduction}
Learning to act from raw, high-dimensional observations like images is arguably the only viable way to scale sensorimotor control to complex real-world setups. However, the main challenge lies in figuring out the representation of these observations upon which the agent's policy is to be learned. The conventional wisdom today is to learn these representations either via task supervision in an end-to-end manner~\cite{dqn,lillicrap2015continuous} or via auxiliary loss functions~\cite{jaderberg2016reinforcement,pathakICMl17curiosity,laskin_srinivas2020curl}. Although such representation learning paradigms are successful in vision~\cite{krizhevsky2012imagenet}, speech~\cite{oord2016wavenet} or language~\cite{devlin2018bert}, they suffer from two key issues in case of robot learning. First, the representations tend to become specific to both the task and the visual environment, thus, any change in the visual input affects the generalization of the learned representations as well as policy. Second, these latent features do not capture the fine-grained location and orientation of objects or robot joints which is indispensable for robotic control in complex scenarios. This is in contrast to sensorimotor representations in humans which have explicit notions of objects, their relationships, 3D reasoning, and geometry~\cite{spelke2007core}.

\begin{figure}[t]
  \centering
  \includegraphics[width=0.48\textwidth]{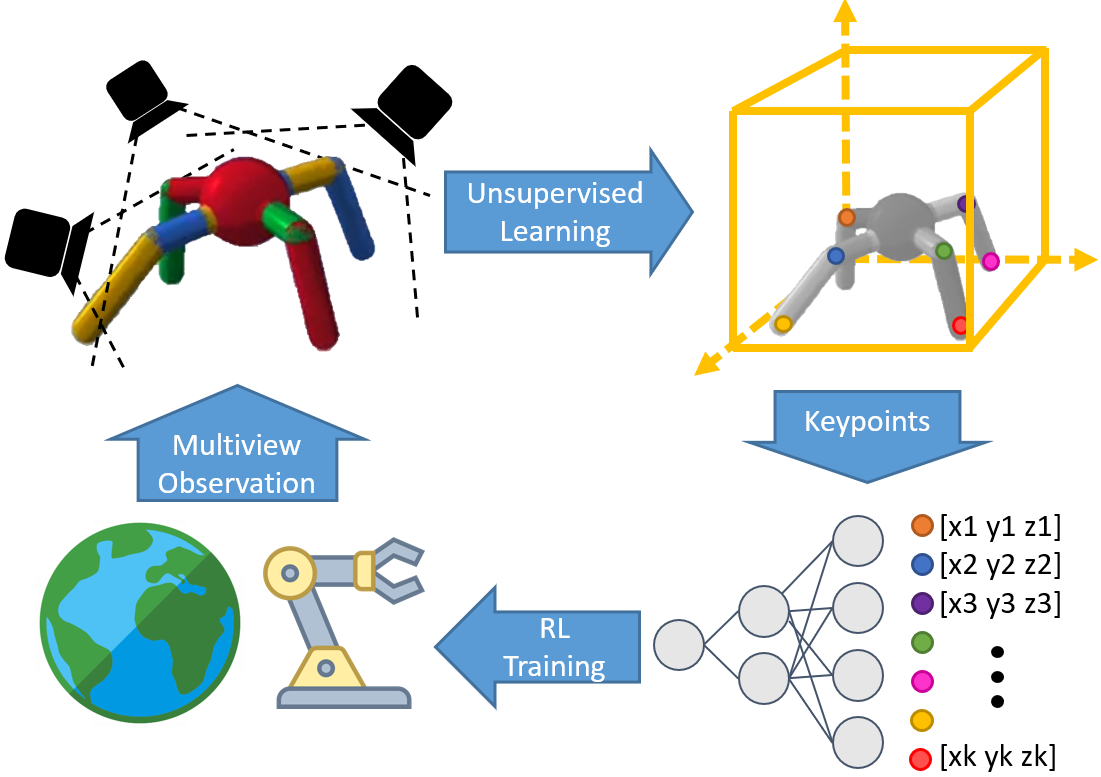}
  \vspace{-0.3in}
  \caption{We propose an end-to-end framework for unsupervised learning of 3D keypoints from multi-view images. These keypoints are discovered to jointly optimize both multi-view reconstruction and downstream task performance. Our learned keypoints are consistent across 3D space as well as time.}
  \label{fig:teaser}
  \vspace{-0.1in}
\end{figure}

\begin{figure*}[t]
  \centering
  \includegraphics[width=\textwidth]{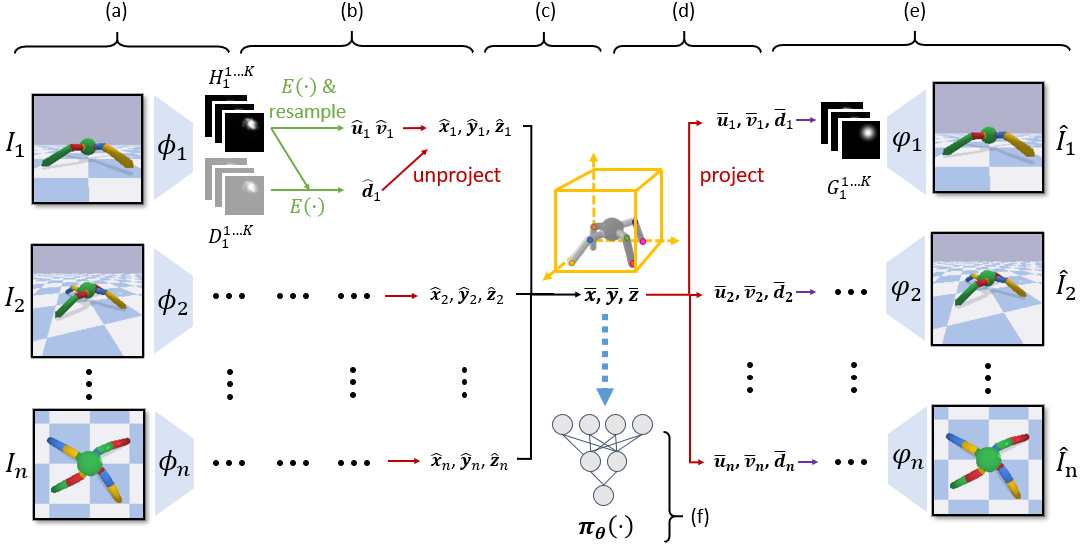}
  \vspace{-0.25in}
  \caption{Overview of our \textit{Keypoint3D} algorithm. (a) For each camera view, a fully convolutional neural network encodes the input image into K heat maps and depth maps. (b) We then treat these heat maps as probabilities to compute expectation of spatial $uv$ coordinates in camera plane. These expected values and the saptial variances are used to resample final $uv$ keypoint coordinates which adds noise that prevents the decoder from cheating to hide the input information in the relative locations $uv$ keypoints. We also take expectation of depth coordinate, $d$, using the same probability distribution. These $[u,v,d]$ coordinates are then unprojected into the world coordinate. (c) We take attention-weighted average of keypoint estimations from different camera views to get a single prediction in the world coordinate. (d) For decoding, we project predicted keypoints in world coordinate to $[u, v, d]$ in each camera plane. (e) Each keypoint coordinate is mapped to a gaussian map, where a 2D gussian is created with mean at $[u, v]$ and std inversely proportional to $d$. For each camera, gaussian maps are stacked together and passed into decoder to reconstruct observed pixels from the camera. (f) Together with reconstruction, we also jointly train a task MLP policy on top of predicted world coordinates via reinforcement learning.}
  \label{fig:method}
\end{figure*}

Motivated by the need to capture fine-grained abstractions from visual data, a popular approach is to use keypoint-based representations. Keypoints are represented in Euclidean coordinates and provide a natural way to represent the kinematic structure of both the agent and the environment. Furthermore, keypoint- or particle-based representation provides an ideal way to represent deformable objects like cloth for precise robotic manipulation~\cite{clegg2017learning,sundaresan2020learning} but these keypoints are often hand-selected by human labelers which is expensive and hinders scalability. Some prior methods try to alleviate this problem~\cite{minderer2019unsupervised,kulkarni2019unsupervised} by learning keypoints in an unsupervised and task-agnostic manner, however, the keypoints are defined in the image coordinate space in 2D. This limits the agent's ability to perform 3D reasoning and occlusion handling which is vital in partially observable or even fully observable environments as the tasks are performed in a 3D world.

An alternative approach to capture 3D is to leverage structure from motion~\cite{wei2020deepsfm} to aggregate multi-view information. Recent works~\cite{manuelli2019kpam,manuelli2020keypoints} learn keypoints by solving for 3D correspondence across views via multi-view stereo but the keypoint discovery phase is performed separately from control and not learned in an end-to-end manner. However, different tasks may demand different parts of the scene to be keypoints. For example, if a robot is to pick up a cup, its handle would be a keypoint but if something is to be poured into it, 3D keypoints should capture its interior circumference. How can we have the best of all worlds: fine-grained keypoints, 3D reasoning, and unsupervised, end-to-end, joint training with control?

In this work, we propose an end-to-end framework for unsupervised discovery of 3D keypoints from images that are learned directly via the task performance, shown in \cref{fig:teaser}. For such an approach to be general as well as successful, it should satisfy three properties: (a) \textbf{Consistency across 3D space}: the learned keypoints should capture the same 3D location in the world coordinate from different views of the same scene. (b) \textbf{Consistency across time}: the same keypoint should track the corresponding entity consistently across different timesteps of trajectory rollout. (c) \textbf{Joint learning with control}: 3D keypoints should be learned such that they are directly useful for the end-task performance. We conceptually integrate these properties within a common mathematical framework.

Given an input image from a camera view, we first predict keypoint locations and depth in the image space. These keypoints from different views of the same scene are then aggregated using camera information to obtain a global world coordinate of each keypoint via a differentiable unprojection operation. The consistency across views is learned via multi-view consistency loss that ensures the keypoint from different view maps to common world coordinate. This global coordinate is then projected back into the image plane of each camera view to reconstruct the original input image of the same view. This setup creates a differentiable 3D keypoint bottleneck upon which the agent's control policy is trained via reinforcement learning. The keypoint extraction and agent's policy are optimized jointly in an end-to-end manner as shown in \cref{fig:method}. We refer to our unsupervised algorithm as \textit{Keypoint3D} in short.

We evaluate Keypoint3D across a variety of reinforcement learning benchmark environments, and we perform the following analyses.  We first investigate how well our Keypoint3D representations perform compared to other representations for RL. Second, we test the scalability to higher dimensional control problems. Third, we show that our Keypoint3D based policy is capable of manipulating deformable objects as evident from results on a task where a robot must put a scarf around a human mannequin. Finally, we show that our Keypoint3D representations generalize across tasks as well. Our method outperforms prior state-of-the-art across almost all the environments and our ablation study demonstrates its robustness across several design choices.

\section{Unsupervised Learning of 3D Keypoints}
We use a multi-view encoder-decoder architecture to learn 3D keypoints without supervision. Given $N$ cameras from different views of a same scene, we associate an encoder and a decoder with each camera. We provide three distinct sources of \textit{unsupervised} training signal for learning of the Keypoint3D representation:  (1) We force the predicted keypoints across encoders to be geometrically consistent in 3D by encouraging the different view-specific keypoint coordinates to unproject to the same point in 3D space.  (2) We impose an image reconstruction loss, which penalizes inaccurate reconstructions from the decoder.  (3) We use the reward incurred by the RL policy that takes as input the learned keypoints, and backpropagate the reward signal through the weights of the encoder.

% \vspace{-0.15in}
\subsection{Preliminaries}
Let $I_{n}\in \mathbb{R}^{H \times W\times C}$ be the image observation from camera $n\in 1...N$ with extrinsic matrix $V_{n}$ and intrinsic matrix $P_{n}$. Let K be the number of keypoints we intend to detect. Keypoints are indexed with $k=1...K$. For a 3D point $[x, y, z]^\top$ in world coordinate, we can use extrinsic matrix $V_n$ and perspective intrinsic matrix $P_n$ to project it to camera coordinate $[u, v, d]^\top$ for camera $n$, where $u,v \in [0,1]$ is a normalized coordinate on camera plane and $d>0$ is the depth value of that point from camera plane. Let the operator $\Omega_{n}: [x,y,z]^\top\to[u,v,d]^\top$ denote this projection, and let its inverse $\Omega_{n}^{-1}: [u,v,d]^\top\to[x,y,z]^\top$ denote the unprojection operator that maps a camera coordinate to a world coordinate. Both $\Omega_{n}$ and $\Omega_{n}^{-1}$ are differentiable and can be expressed analytically.

% \paragraph{Keypoint Detector}
\subsection{Differentiable 3D Keypoint Bottleneck}
We want to leverage the spatial inductive bias of our fully convolutional auto-encoder yet ensure the latent \text{learned latent representation} is in the form of xyz coordinates rather than feature maps. To achieve this, we need a differentiable keypoint bottleneck that maps from dense maps to sparse coordinates. While a extraction for world coordinates is not possible from encoded feature space directly, \cite{jakab2018unsupervised} offers a way to parameterize points in uv coordinates on camera plane using spatial probability. We can add depth prediction to the uv coordinate prediction and extract world coordinate using projection geometry. 

\paragraph{Keypoint Detector} For each camera $n$, we pass $I_{n}$ into a fully convolutional encoder $\phi_{n}$ to get $K$ confidence maps $C_{n}^{k}\in \mathbb{R}^{S \times S}, k=1... K$ and depth maps $D_{n}^{k}\in \mathbb{R}^{S \times S}, k=1... K$. For each confidence map $C_{n}^{k}$,  we can take a spatial softmax to compute a probability heatmap $H_{n}^{k}$:
$H_{n}^{k}(i,j)=\frac{\exp(C_{n}^{k}(i,j))}{\sum_{p=1}^{S}\sum_{q=1}^{S}\exp(C_{n}^{k}(p,q))}$.
Each entry in heatmap $H_{n}^{k}\in \mathbb{R}^{S \times S}$ represents the probability of a 3D keypoint $k$ to be at that position on the 2D image plane if viewed from camera $n$. Each depth map $D_{n}^{k}$ is a dense prediction of distance from camera plane for 3D keypoint $k$ being at each position.

For each pair of heatmap $H_{n}^{k}$ and depth map $D_{n}^{k}$, we can extract the expected 3D position of the $k$-th keypoint in the $n_{th}$ camera's coordinate. We can calculate the expected $[u,v,d]$ camera coordinate over the probability map using the following equations. Note that crucially, we are taking an expectation over the map of \textit{coordinates}, with weights given by the predicted heatmap values.
\begin{align*}
\mathbb{E}[u_{n}^{k}]&=\frac{1}{S}\sum_{u,v}u\cdot H_{n}^{k}(u,v); \quad
\mathbb{E}[v_{n}^{k}]=\frac{1}{S}\sum_{u,v}v\cdot H_{n}^{k}(u,v)\\
\mathbb{E}[d_{n}^{k}]&=\sum_{u=1}^{S}\sum_{v=1}^{S}D_{n}^{k}(u,v)\cdot H_{n}^{k}(u,v)
\end{align*}

Let $[\hat{u}_{n}^{k}, \hat{v}_{n}^{k}, \hat{d}_{n}^{k}]^\top=[\mathbb{E}[u_{n}^{k}],\mathbb{E}[v_{n}^{k}],\mathbb{E}[d_{n}^{k}]]^\top$.
We found one way for the system to not capture meaningful 3D keypoints is cheating to hide the information about input image in the relative locations keypoint to each other. To avoid this issue, we do not use the exact keypoint locations predicted by the encoder but resample them from by assuming a gaussian distribution around the mean keypoint with standard deviation computed via spatially across the heatmap. 
This adds some stochasticity in the exact location of keypoint coordinates preventing the collapse. More details of this resampling process are in the appendix.

\paragraph{Attention}
\label{sec:attention}
After predicting keypoints in each of our $n$ camera coordinate frames, we need a way to combine the $n$ coordinates into one.  A naive approach is to simply average the predictions from each view.  However, certain keypoints might be occluded from certain view points, and might thus get predicted poorly. To address this, we repurpose our predicted confidence maps in order to compute weights for a weighted average. This allows us to ignore the less confident keypoint estimations which may hurt the task performance. 

Recall that the $(i, j)$ entry of the confidence map $C_{n}^k$ denotes the predicted log likelihood of keypoint $k$ appearing at pixel $(i, j)$ in the image from camera $n$.  We then assign a ``confidence score" for the $n$-th encoder's prediction for keypoint $k$.  This score is proportional to the mean of the confidence map $C_{n}^k$, and the scores of the $k$ keypoints are normalized to sum to 1 via a softmax:
\[A_{n}^{k}=\frac{\exp(\frac{1}{S^2}\sum_{p=1}^{S}\sum_{q=1}^{S}C_{n}^{k}(p,q))}{\sum_{i=1}^{K}\exp(\frac{1}{S^2}\sum_{p=1}^{S}\sum_{q=1}^{S}C_{n}^{i}(p,q))}\]

\paragraph{Extracting world coordinates}
Given predicted keypoints $[\hat{u}_{n}^{k}, \hat{v}_{n}^{k}, \hat{d}_{n}^{k}]^\top, n=1...N, k=1...K$ in camera coordinats, we can then unproject them to world coordinates using $[\hat{x}_{n}^{k},\hat{y}_{n}^{k},\hat{z}_{n}^{k}]^\top=\Omega_{n}^{-1}([\hat{u}_{n}^{k}, \hat{v}_{n}^{k}, \hat{d}_{n}^{k}]^\top)$. This is the predicted world coordinate of the $k^{th}$ keypoint conditioned on the image from camera $n$. For each keypoint, we have $N$ independent predictions from $N$ cameras. We then take an average over these predictions for each keypoint weighted with normalized confidence, that is \[[\bar{x}^{k},\bar{y}^{k},\bar{z}^{k}]^\top=\sum_{n=1}^{N}\frac{A_{n}^{k}}{\sum_{m=1}^{N}A_{m}^{k}}\cdot[\hat{x}_{n}^{k},\hat{y}_{n}^{k},\hat{z}_{n}^{k}]^\top\]

\paragraph{Keypoint Decoder}
To also leverage spatial inductive bias for the decoder, we need to give keypoint coordinates spatial structure again before passing them into the decoder. This can be achieved by reprojecting 3D keypoints to each image view again and constructing a 2D gaussian for each keypoint on the camera plane. We reproject all $K$ keypoints in world coordinate to the $N$ camera planes to regain spatial structure before decoding. Using the projection operator, for each camera and keypoint we can get $[\bar{u}, \bar{v}, \bar{d}]^\top=\Omega_{n}([\bar{x}, \bar{y}, \bar{z}]^\top)$. To regain spatial structure, for each camera and keypoint we create Gaussian maps $G_{n}^{k}\in \mathbb{R}^{S\times S}$. Each Gaussian map is a 2D gaussian with mean $[\bar{u}, \bar{v}]^\top$ and covariance matrix $\mathbf{I}_2 / \bar{d}$, where $\mathbf{I}_2$ is the $2\times 2$ identity matrix. This matrix makes closer point have a larger spatial span in the gaussian map corresponding to that camera. Let us define averaged attention across all views as $\bar{A}^{k} = \frac{1}{N}\sum_{n=1}^{N}A_{n}^{k}$. The decoder $\psi_{n}$ for each camera takes the stacked Gaussian maps $G_{n}$ to predict $I_{n}$ where $G_{n}=K\cdot \text{stack}([G_{n}^{1}\bar{A}_{n}^{1},G_{n}^{2}\bar{A}_{n}^{2}...G_{n}^{K}\bar{A}_{n}^{K}])$. The decoder $\psi_{n}$ then decodes the stacked Gaussian maps to reconstruct the observed image.

\begin{algorithm}[t]
    \caption{Keypoint3D: RL with 3D Keypoint Bottleneck}
    \label{alg:rlalgo}
    \begin{algorithmic}
       \STATE {} 
       \textcolor{algo_color}{
       \STATE $B_{obs} \gets \text{\O}$
       }
       \FOR{$update=1, 2...$}
           \STATE $B_{rollout} \gets \text{\O}$
           \FOR{$actor=1, 2...,N$}
               \item Run policy $\pi_{\theta_{old}}$ for $T$ steps to get $(s,a,r)_{1...T}$
               \textcolor{algo_color}{
               \STATE $B_{obs} \gets  s_{1...T}\cup B_{obs}$
               }
               \STATE $B_{rollout} \gets (s,a,r)_{1...T} \cup B_{rollout}$
           \ENDFOR
           \textcolor{algo_color}{
           \FOR{$i=1, 2... p$}
               \item Optimize $L_{unsup}$ with $s\sim B_{obs}$ wrt $\theta_{ae}$
           \ENDFOR
           }
           \item Compute advantage estimates $\hat{A}_{1:N,1:T}$ for $B_{rollout}$
           \FOR{$epoch=1, 2... q$}
               \item Optimize $L_{policy}+\textcolor{algo_color}{L_{unsup}}$ throughout $B_{rollout}$
           \ENDFOR
           \STATE $\theta_{old} \gets \theta$
       \ENDFOR
    \end{algorithmic}
\end{algorithm}

\subsection{Losses for Training 3D Keypoint Encoder}
Our Keypoint3D approach jointly optimizes the multi-view reconstruction via unsupervised learning and the task policy via reinforcement learning. We backpropagate the sum of unsupervised learning loss and policy loss to train the Keypoint3D pipeline. The unsupervised losses have three components as defined below. 

\paragraph{Multi-view Auto Encoding Loss} We encourage the keypoints to track important entities and structures. An Auto-encoding loss has been shown to be useful for this, as the architectural bottleneck forces the latent representation to capture the most salient aspects of the scene: $L_{ae}=\sum_{n=1}^{N}||\psi_{n}(G_{n})-I_{n}||_{2}$.

\begin{figure*}[t]
\centering
\includegraphics[width=\linewidth]{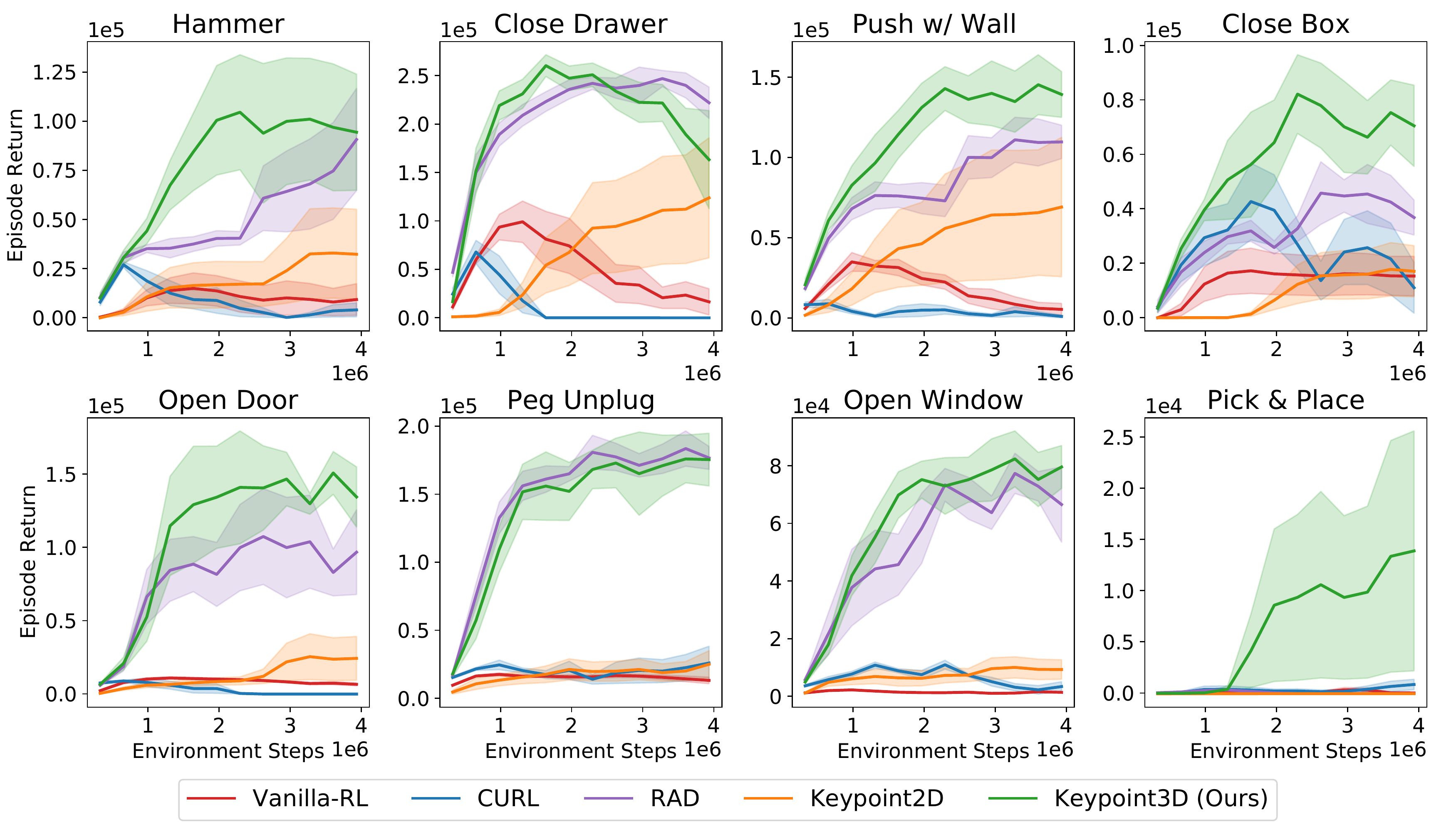}
\vspace{-0.33in}
\caption{Plots show the performance of \textit{Keypoint3D} on 8 metaworld environments with varying difficulty compared to different representation learning methods. Mean and standard error are shown across 4 random seeds for each environment. Our 3D keypoint method outperforms the strongest baseline, RAD, on 5 out of the 8 environments while being similiar in the rest.}
\label{fig:metaworld}
\end{figure*}

\paragraph{Multi-view Consistency Loss}
We enforce that for each keypoint $k$, the camera-frame coordinates of the keypoint $k$ from each of the $n$ encoders all have identical 3D world coordinates. We use a multi-view consistency loss to penalize disagreement between predictions from different views:
\vspace{-0.1in}
\[L_{multi}=\sum_{k=1}^{K}\sum_{i=1}^{N}\sum_{j=1}^{N}||[\hat{x}_{i}^{k},\hat{y}_{i}^{k},\hat{z}_{i}^{k}]^\top-sg([\hat{x}_{j}^{k},\hat{y}_{j}^{k},\hat{z}_{j}^{k}]^\top)||_{2}\] where $sg$ is the stop gradient operator.

\paragraph{Seperation Loss}
Finally, the keypoints should not collapse together and should keep a distance from each other to effectively track different objects in the scene: 
\[L_{sep}=\frac{1}{K^2}\sum_{i=1}^{K}\sum_{j=1}^{K}\frac{1}{1+M*||[\bar{x}^{i},\bar{y}^{i},\bar{z}^{i}]^\top-[\bar{x}^{j},\bar{y}^{j},\bar{z}^{j}]^\top||_{2}^2}\] 
where M is a positive value chosen to be $1000$.

\noindent Finally, the total unsupervised learning loss is:
\vspace{-0.1in}
\[L_{unsup}=\lambda_{ae}\cdot L_{ae}+\lambda_{multi}\cdot L_{multi}+\lambda_{sep}\cdot L_{sep}\]

\subsection{Joint Learning of 3D Keypoints with Control}
\label{sec:rlmethod}
We integrate our 3D keypoint learning with PPO~\cite{DBLP:journals/corr/SchulmanWDRK17} so a control policy can be trained jointly on top of learned keypoints. We keep an observation buffer for all observed images. Before each policy gradient update, we train unsupervised learning from observations sampled from this buffer for several steps, updating encoder and decoder parameters $\theta_{ae}$ with $L_{unsup}$. During policy gradient update, we optimize all parameters with the sum of policy loss $L_{policy}$ and unsupervised learning loss $L_{unsup}$. The policy is learned on top of keypoints represented in world coordinates which can have arbitrary scaling and can hurt the training process. To handle this, we \textit{renormalize} the values of predicted 3D keypoints before feeding them to actor and critic network. Pseudo code for the method is provided in \cref{alg:rlalgo} with our addition in \textcolor{algo_color}{magenta} color.

\paragraph{Temporal Consistency of 3D Keypoints}
\label{sec:temporalvariant}
In many pixel-observation environments that requires temporal reasoning, a common technique is stacking adjacent frames as observation~\cite{mnih2015human}. In our method, we use the technique but also apply the same keypoint detector on each frame independently with shared encoder weights. The result is thus stacked keypoints across frames. We compute a difference vector indicating keypoint movements across adjacent frames and concatenate it with latest keypoints before feeding into fc layers of policy network. Note we further add data augmentation which forces the keypoint to be consistent across different appearances, thereby granting consistency over time. We also tried adding a temporal prediction loss but it didn't help because of sufficient temporal tracking signal from the data augmentations. 

\subsection{Data Augmentation as a Self-Supervisory Signal}
\label{sec:augmentation}
Because the keypoint uv coordinates have explicit geometric interpretations, we can leverage data augmentation~\cite{krizhevsky2012imagenet} to provide additional self-supervision. The core intuition is that, when the input image is translated, the predicted keypoints, projected on camera plane, should also get translated by the same amount. Because the random shift operator, $f$, effectively transforms heatmap coordinates from the original unshifted image, we apply a reverse transformation, $f^{-1}$,  on the $\mathbb{E}[u_{n}^{k}]$, $\mathbb{E}[v_{n}^{k}]$ before unprojection to world coordinate. Before constructing gaussian maps, $f$ is applied to reprojected keypoints again to map them back to the shifted coordinate. The random translation with shift offset plays a critical role in getting high quality keypoints.

\section{Experimental Setup}
We evaluate our method on a variety of 3D control tasks with three-camera pixel observation. A reinforcement learning policy is trained jointly with unsupervised 3D keypoint learning as described in \cref{sec:rlmethod}. We choose a set of 3D manipulation environments~\cite{yu2019metaworld}, a high-dof 3D locomotion environment~\cite{coumans2019}, a customized soft-body environment and a meta-learning benchmark~\cite{yu2019metaworld} to evaluate our method from different perspectives. These environments are originally developed for state based RL and are hard tasks for pixel based RL. We setup three third-person-view cameras in the scene such that most objects are visible from all three views. More details in the appendix.

We compare \textit{Keypoint3D} with a variety of visual reinforcement learning algorithms. All the baselines, including Keypoint3D, are implemented on top of PPO~\cite{DBLP:journals/corr/SchulmanWDRK17} with the same CNN architecture and algorithm hyper-parameters. We stack images from all three view as observation for the baselines: (1) RAD~\cite{laskin_lee2020rad} is the state-of-art method for pixel based reinforcement learning. It achieves high sample efficiency through data augmentation. (2) CURL~\cite{laskin_srinivas2020curl} trains reinforcement learning on top of learned representation through contrastive learning. (3) Vanilla PPO is the original PPO algorithm \cite{DBLP:journals/corr/SchulmanWDRK17} with a CNN architecture to take in image observations. (4) Keypoint 2D is PPO-based implementation of~\cite{minderer2019unsupervised}. We choose the number of 2D keypoint to be $3/2$ of that of 3D keypoints so both methods have same number of coordinate bits. 

\section{Results and Analysis}
We investigate following questions: a) How well does our 3D keypoint representation compare to other representations for policy learning? b) How well does our 3D keypoint method scale to high dimensional control from high dimensional observations? c) How well does our method adapt to deformable object manipulation where keypoints are more desirable but harder to track? d) How well does our 3D keypoint representation generalize to unseen setups?

\subsection{Accuracy and Efficieny of 3D Keypoints}
\label{sec:metaworldexp}
We select 8 tasks of varying difficulty from meta-world, each featuring 50 random configurations. As shown in \cref{fig:metaworld}, our method out performs the state-of-art method RAD by a margin in 5 out of the 8 environments while having similar performance in rest of the environments.

CURL and vanilla PPO performed poorly on all 8 visual metaworld environments. Vanilla PPO and RAD usually perform well in easier tasks where a camera attached to end effector will suffice. Such setups assume a strong inductive bias that a big change in task space, such as reward, is associated with large pixel space change, such as an object that occupies half of the camera image. In a third person view setup such inductive bias is no longer true and a change in task space can correspond to an insignificant change in pixel space, if at a distance from camera. RAD mitigates this problem by using data augmentation to let encoder focus on a larger set of global features yet still doesn't explicitly distill structure of the scene. In contrast, our method uses keypoints to distill the fine-grained movement of all moving components in the scene. 2D Keypoint performs better than CURL and Vanilla PPO but didn't outperform our 3D Keypoint method.
This is likely because, in metaworld environments, objects can get very close to camera and show a large surface, which can be a problem for the 2D Keypoint representation ~\cite{minderer2019unsupervised} as it uses a fixed spatial variance in Gaussian map. Our 3D Keypoint method significantly mitigates the problem by assigning large spatial variance in gaussian map for closer objects, telling the decoder the presence of a potential large surface. 

\begin{figure}[t]
\centering
\includegraphics[width=\linewidth]{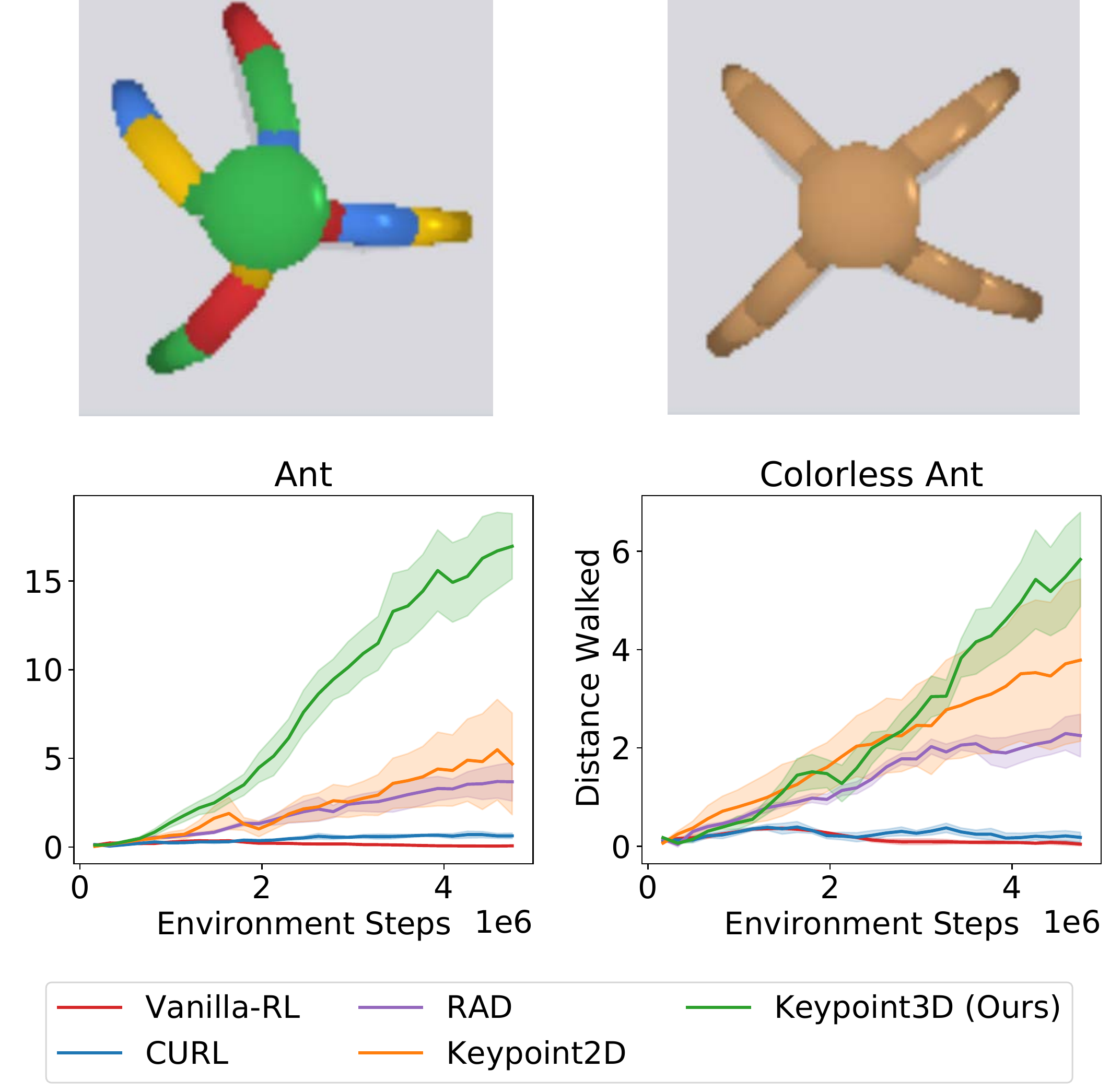}
\vspace{-0.25in}
\caption{Benchmark reinforcement learning on Pybullet Ant (Left), a highly dynamical environment where fine-grained understanding of movable joints and parts are essential. Colorless Ant(Right) is a modified version without homogeneous limb colors to evaluate our method in low-texture setting.}
\label{fig:ant}
\vspace{-0.1in}
\end{figure}

\subsection{Scaling to High-Dimensional Control}
\label{sec:antexp}
To evaluate our method on high dimensional control, we choose a highly dynamical 3D-locomotion environment, Pybullet \cite{coumans2019} Ant, where fine-grained understanding of movable joints and parts are essential. As locomotion environments require temporal reasoning, we use a frame stack of 2. The original ant environment in Pybullet assigns different colors to adjacent limbs. To further show that our method is robust to low-textured objects, we also benchmark on a colorless variant of ant.

\begin{figure*}[t!]
\centering
\includegraphics[width=0.95\linewidth]{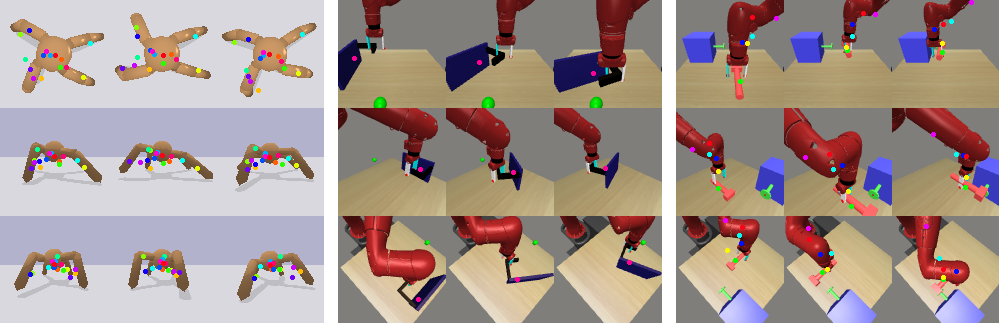}
\vspace{-0.1in}
\caption{Visualization of the learned keypoints on colorless ant (left), metaworld close door (middle), metaworld hammer (right) environments. For the two metaworld environments, we filter keypoints with a threshold on learned keypoint attention. In colorless ant, the keypoints consistently track the limbs. In door the pink point with highest confidence tracks the movement of the door consistently. In hammer, the green point tracks hammer, while other points tracks the end-effector and sections of the arm respectively.}
\label{fig:visualization}
\vspace{-0.1in}
\end{figure*}

As shown in \cref{fig:ant}, our method significantly outperforms all baselines in both ant and its colorless variant. Our method, in particular, avoids local minimum as it better captures the structure of the robot as visualized in \cref{fig:visualization}, where the Keypoint3D captures the 3D movement of the ant robot in world coordinates. The ant environment, from pixels, requires the temporal knowledge of velocity of body and limb to apply control effectively. We believe the temporal consistency as described in \cref{sec:temporalvariant} exactly achieved this, estimating the movement vector and feeding the mission-critical velocity information into our policy just like in state-space. Superiority of keypoint-based method is also reflected in the high performance of 2D keypoint baseline on the environment. On the other hand, Vanilla-RL, RAD and CURL lack explicit modeling of movement. CURL's contrastive learning will overly focus on pixels and fails to capture the fine-grained joint positions or movements that results in little pixel change.

\begin{figure}[t]
\centering
\includegraphics[width=0.95\linewidth]{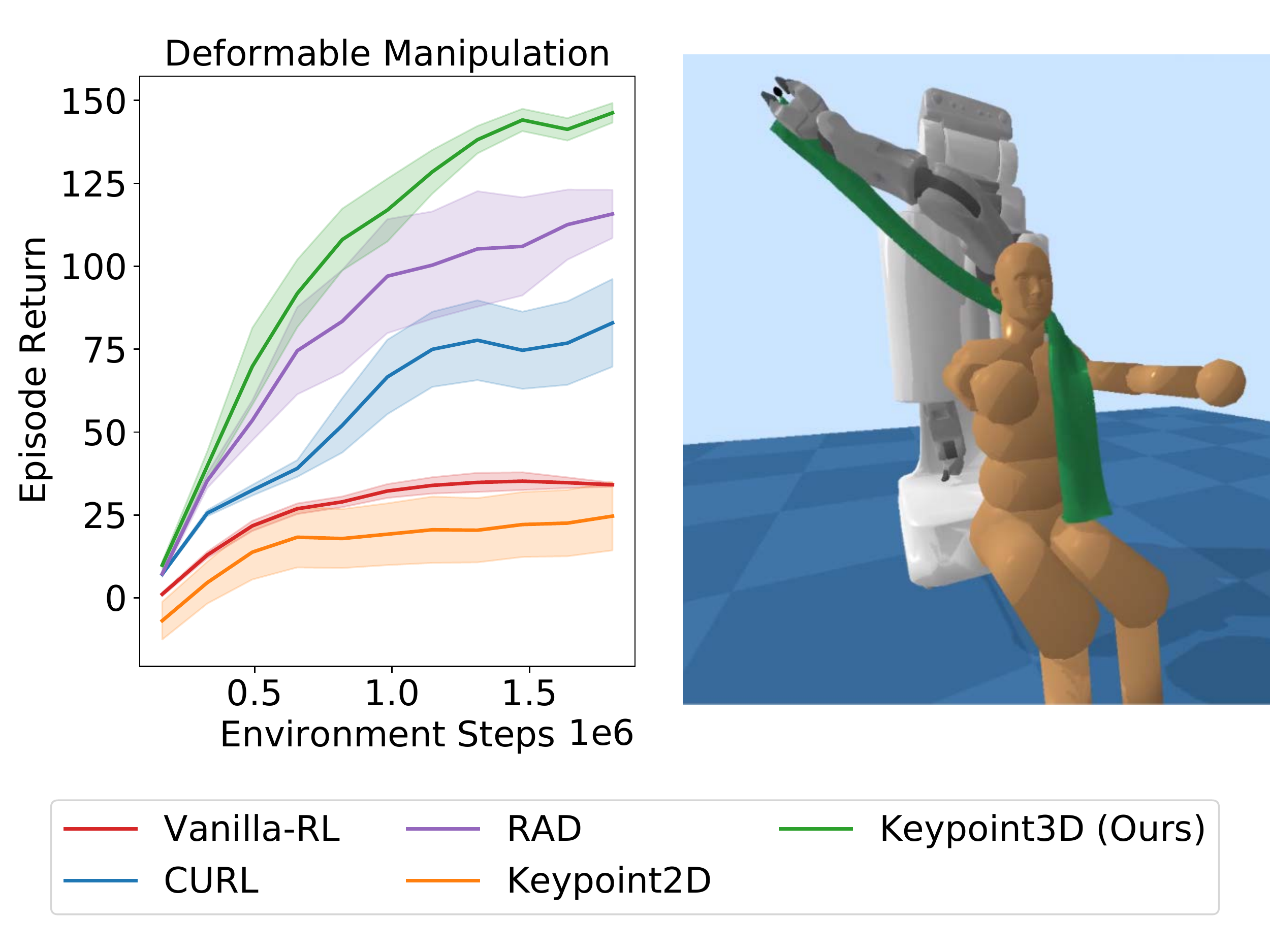}
\vspace{-0.2in}
\caption{Left: Learning curve on scarf manipulation environment. Our Keypoint3D approach outperforms all the prior works consistently. Right: Scarf manipulation environment. The objective is to wind the scarf around human neck for one and a half full circle.}
\label{fig:scarf}
\vspace{-0.1in}
\end{figure}

\subsection{Keypoints for Deformable Object Manipulation}
\label{sec:scarfexp}
We further evaluate our method on a customized 3D scarf manipulation environment based on~\cite{erickson2020assistivegym}. In this environment, the objective is putting a scarf around human neck. This task requires both 3D reasoning of occlusion and understanding of the highly dynamical 3D soft object. \cref{fig:scarf} shows the result of Keypoint3D on scarf manipulation is the best among all methods.

\begin{figure}[t]
\centering
\includegraphics[width=\linewidth]{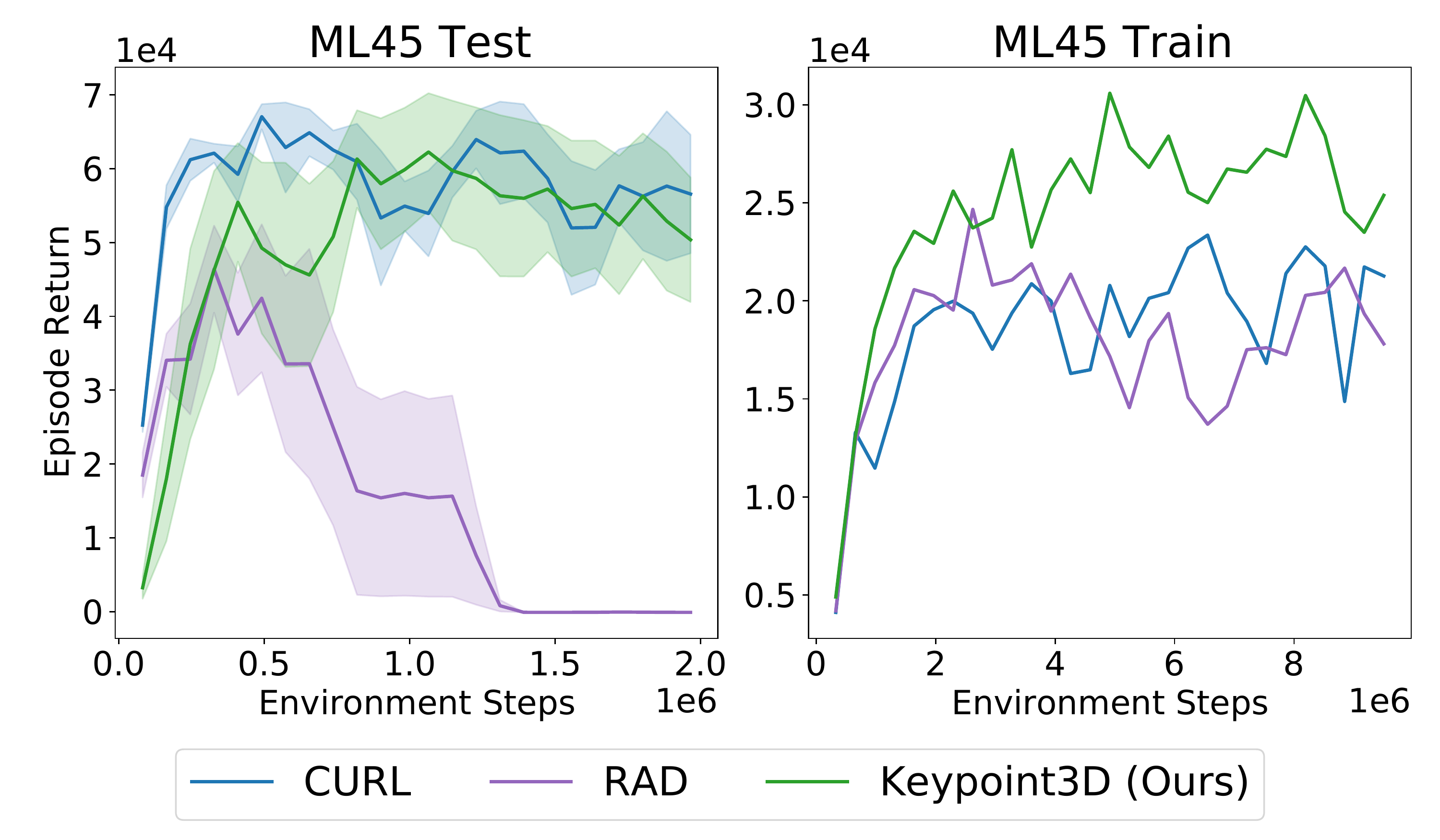}
\vspace{-0.3in}
\caption{Transfer Learning result on ML45 multi-tasking benchmark. Our 3D Keypoint method out performs others by a margin in a training environment consist of 45 tasks. When finetuned on a 5-task test environment, both representation learning methods, CURL and 3D Keypoint, outperforms RAD.}
\label{fig:transfer}
\vspace{-0.2in}
\end{figure}

\subsection{3D Keypoints for Transfer Learning across Tasks}
\label{sec:transfer-learning}
To test how well does our 3D keypoint representation generalize to unseen objects, we conduct a transfer learning experiment on the ml45 multi-task learning benchmark of metaworld \cite{yu2019metaworld}. We followed the train/test split of the benchmark, first pre-train our method and baselines on 45 training environments featuring distinct objects for 10M steps. We then conduct transfer learning on 5 unseen test environments with the pre-trained weights for 2M steps. For each method, we freeze the pre-trained encoder and finetune the trained mlp part of the policy. We compare our method against the transfer result of RAD as well as CURL as shown in \cref{fig:transfer}. Our method outperforms all the baselines in the training environment for the same reason in the single-task setting of metaworld. Both our 3D Keypoint method and CURL shows great transfer result compared to RAD, which fails because it does not have a representation learning component.

\subsection{Visualization of Discovered 3D Keypoints}
\label{sec:visualization}
To better understand how our 3D keypoint learning method capture meaningful 3D structure of the tasks, we visualize the learned keypoints by projecting them onto the camera plane of each view. Keypoints of the same color across different views correspond to the same keypoint in 3D space. We can also filter out a portion of the keypoints with a threshold of learned attention. \cref{fig:visualization} illustrates that the learned keypoints with most attention consistently follow the movement of essential moving components in the scene. In the colorless ant environment, keypoints track the 12 limbs and joints of the ant robot throughout time despite being colorless. The same phenomenon can be observed in the metaworld environments. More visualizations are shown in the appendix.

\begin{figure}[t]
\centering
\includegraphics[width=\linewidth]{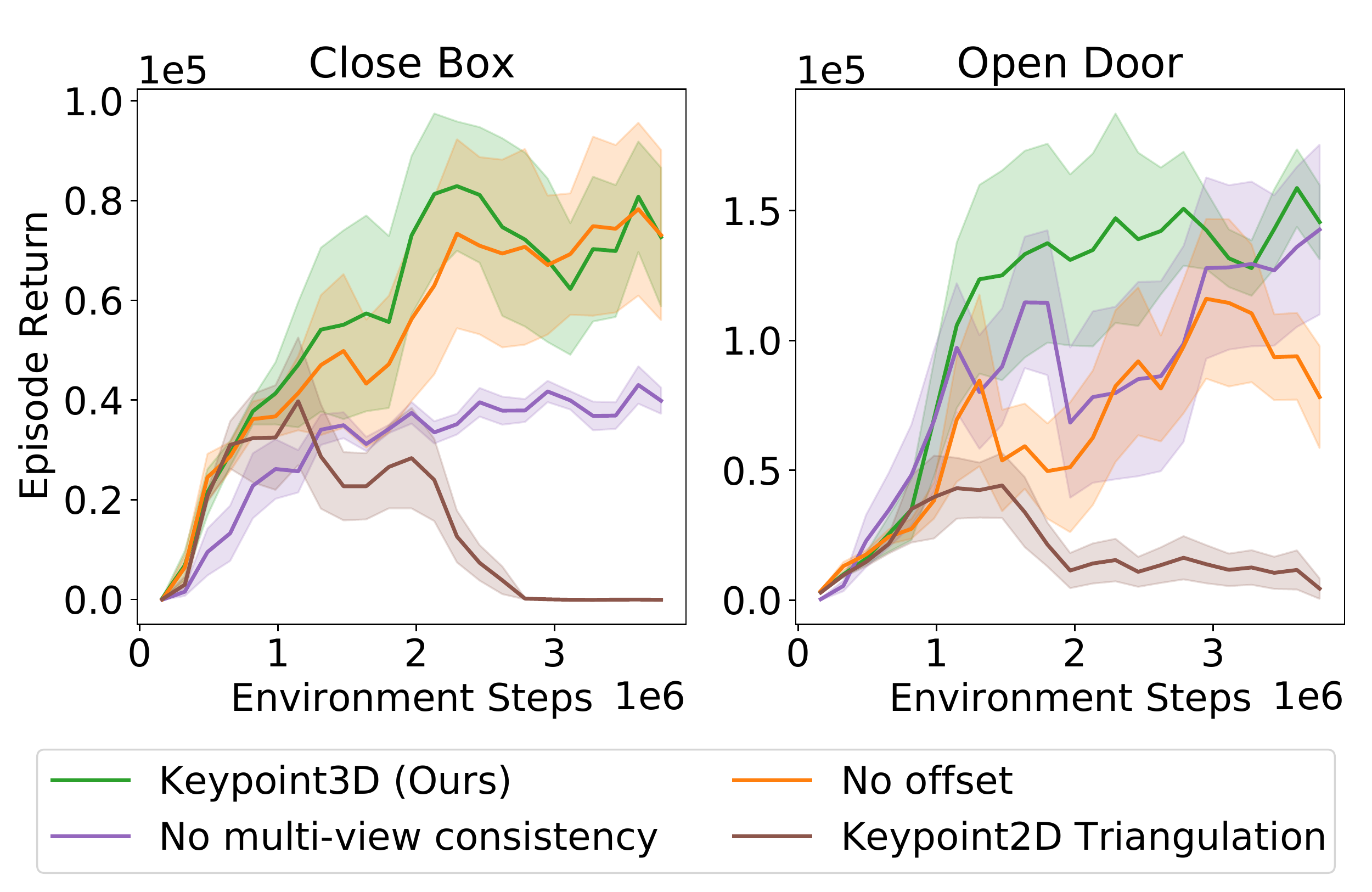}
\vspace{-0.3in}
\caption{Ablations of our Keypoint3D approach with respect to different design choices. Variations include: remove camera offset in the cropping data-augmentation, remove multi-view consistency loss, map keypoints in 2D baseline to 3D space using triangulation.}
\label{fig:ablate1}
\vspace{-0.05in}
\end{figure}

\begin{figure}[t]
\centering
\includegraphics[width=\linewidth]{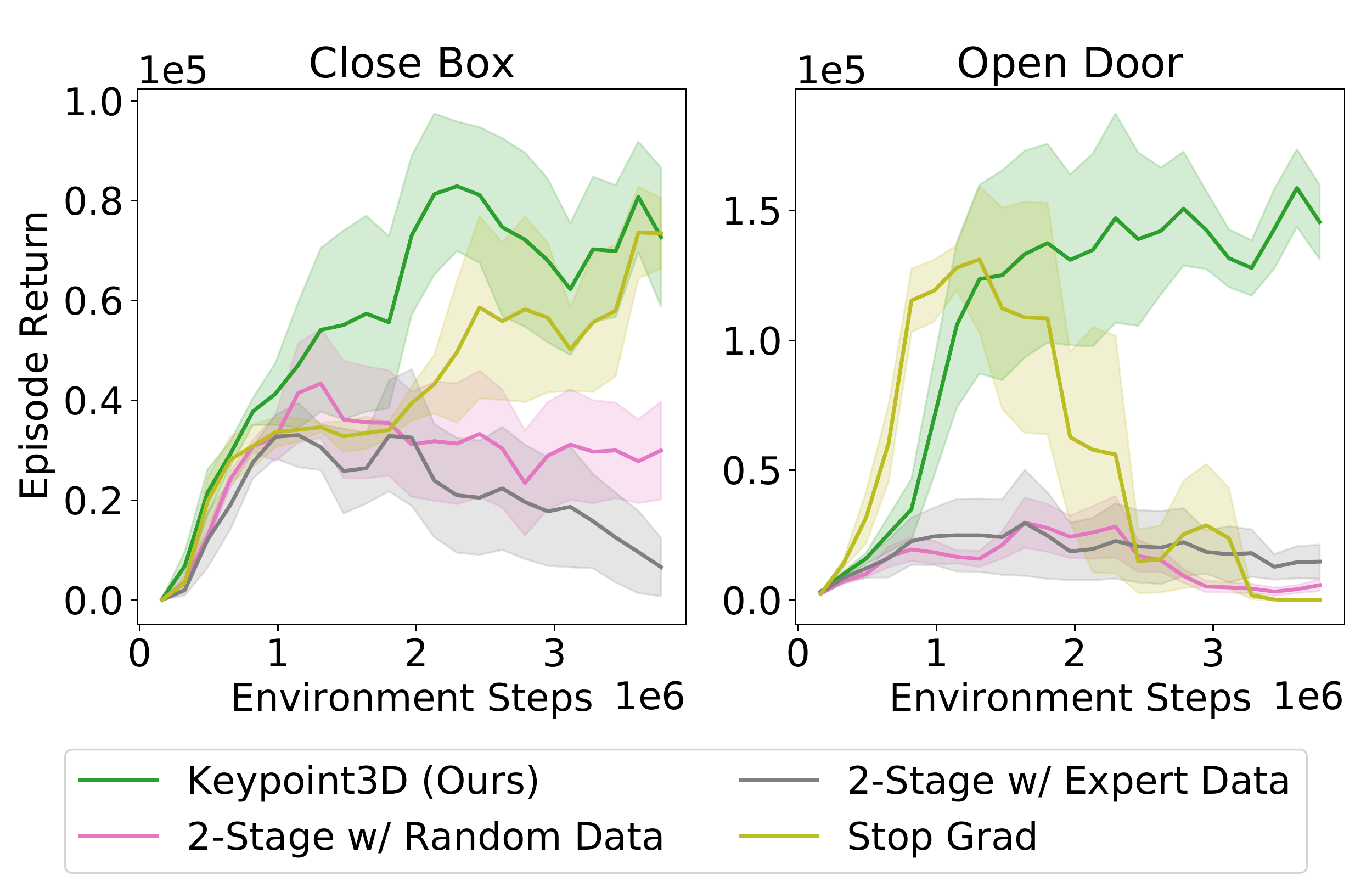}
\vspace{-0.3in}
\caption{Analyzing the role of joint training of unsupervised keypoint learning with control policy. We compare to three different variants of training policy disjointly where keypoint training data is collected via random policy (2-stage w/ Random) and via expert policy (2-stage w/ Expert). Third approach is to keep the same pipeline as ours but stop backpropagation from policy branch. The results show that joint training is crucial.}
\label{fig:ablate2}
\vspace{-0.05in}
\end{figure}

\begin{figure}[t]
\centering
\includegraphics[width=\linewidth]{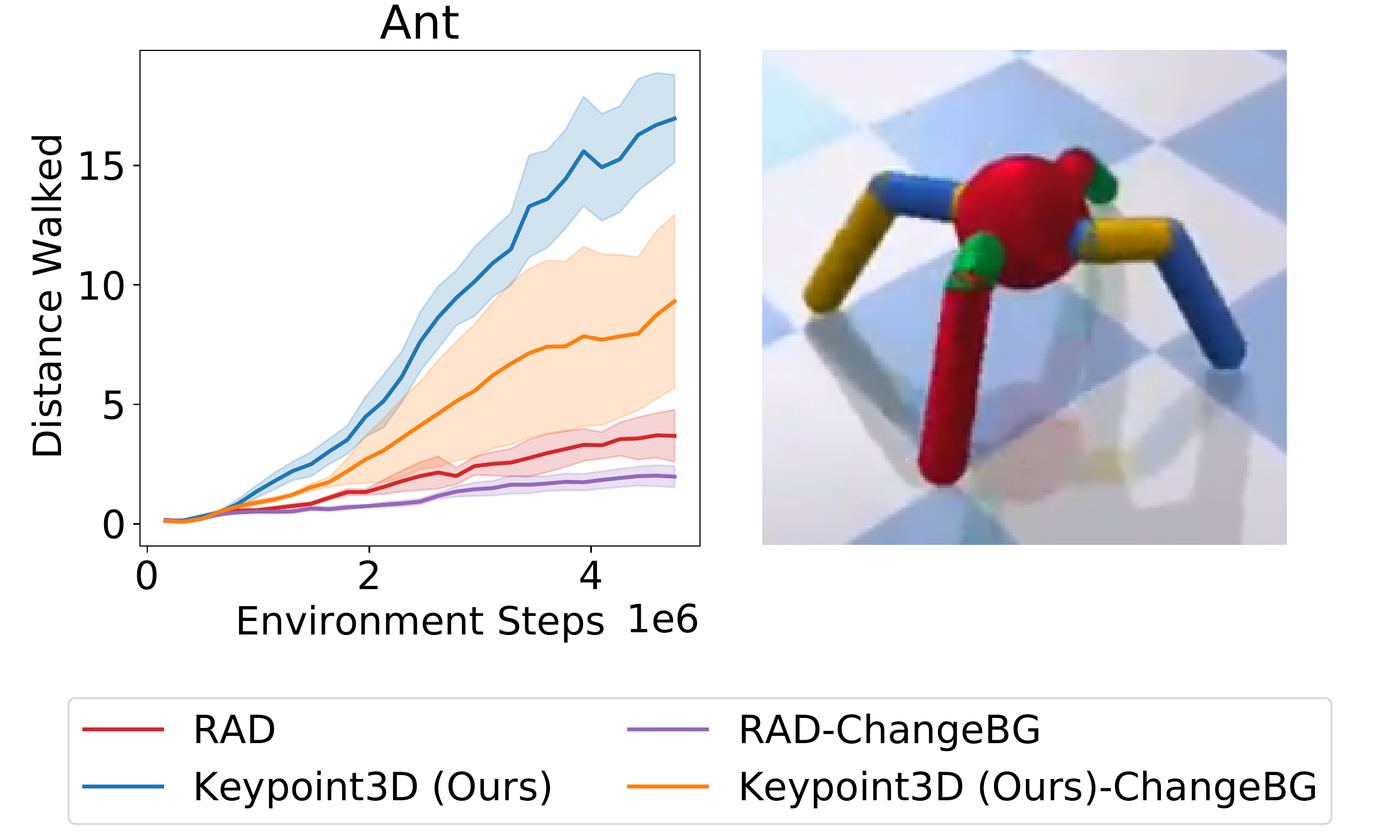}
\vspace{-0.25in}
\caption{Ablation with respect to background changes in high-dimensional control. We show that Keypoint3D is robust to changes in the background unlike other methods because keypoints focus on the agent.}
\label{fig:gridant}
\vspace{-0.12in}
\end{figure}

\section{Analysis and Ablation}
In order to better understand the effect of different components on the sample efficiency of our method, we conduct ablative studies on two metaworld environments. 

\paragraph{Effects of crop offset}
Removing the random crop offset, described in the \cref{sec:augmentation}, significantly lowered our performance on the open door environment as shown in \cref{fig:ablate1}. This is expected as the random shift of the input image should result in a corresponding change in the keypoint coordinate. When the crop offset is removed, randomly shifted keypoints will be incorrectly unprojected as we don't change the camera matrix.

\paragraph{Effects of multi-view consistency}
The ablation results for removing the multi-view consistency loss are shown in \cref{fig:ablate1}. No multi-view consistency severely hurts the performance illustrating that 3D representation is indeed a better representation. We also experiment with a variant of Keypoint2D where we map all 3D keypoints to 3D coordinates using multi-view triangulation. The performace is even worse, indicating multi-view consistency is critical even if we use triangulation instead of depth predictor.

\paragraph{Effects of joint training}
To examine the benefit of training keypoint detection with policy jointly, we carry out three ablations as follows: (a) First train unsupervised 3D keypoint learning on random exploration data. Then train the policy with encoder weights frozen. (b) First train unsupervised 3D keypoint learning on expert data. Then train the policy with encoder weights frozen. c) We stop gradient of policy from backpropagating into encoder. (a)(b) are both 2-stage variants with offline data while (c) examines the effect of joint gradient with on-policy data. Our method outperforms all these three ablations as shown in \cref{fig:ablate2}, showing joint training is important to achieve high sample efficiency as well as performance.

\paragraph{Robustness under changing background}
All of the previous experiments have static background due to the multi-view setup. To evaluate the effectiveness of Keypoint3D under changing background, we run an ant environment variant with checkerboard floor. As shown in \cref{fig:gridant}, changing background of the environment makes the task harder from pixels for all methods. But our method still outperforms the strongest baseline, RAD.

\section{Related Work}
\paragraph{Representation Learning for RL} Common approach in RL is to learn image representations via end-to-end CNN policy~\cite{mnih2013playing,mnih2015human,mnih2016asynchronous}. Recent works enrich this representation via data augmentation~\cite{kostrikov2020image,laskin_lee2020rad}. Alternative to data augmentation is to learn representations via auxiliary losses like inverse model~\cite{pathakICMl17curiosity}, pixel change~\cite{jaderberg2016reinforcement}, VAEs~\cite{kingma2013auto,burda2018large}, or contrastive loss~\cite{laskin_srinivas2020curl}.

\vspace{-0.05in}
\paragraph{Unsupervised Keypoint Learning} Keypoints has been widely used as the representation of structure in image animation \citep{siarohin2020first,siarohin2019animating}, video prediction~\citep{kim2019unsupervised,minderer2019unsupervised} or control~\citep{kulkarni2019unsupervised,minderer2019unsupervised}. These works builds on a fully differentiable keypoint bottlebeck \citep{jakab2018unsupervised} that can map a spatial probability map to images coordinates. They achieve their goal of animating a human or robot in source image by giving a target pose, represented by keypoints, before decoding. However, most of these works learn 2D keypoints in image space. In 3D setting, an object appearing to be a point at distance can be a large chunk of pixels when moving closer to camera. These methods would thus fail when the scene contains 3D movements. 

\vspace{-0.05in}
\paragraph{3D Keypoints Learning}
\citet{DBLP:journals/corr/abs-1807-03146} proposes a method to learn category specific 3D keypoints without supervision given a large dataset of rendered image and camera matrix tuples from many different views of the shapenet \cite{chang2015shapenet} dataset. The method learns category specific 3D keypoints without direct supervision through multi-view consistent pose prediction. It requires this large data set of rendered images from many angles centered around rigid objects. In reinforcement learning setting, however, a scene is dynamic rather than a single rigid object. Cameras are also usually fixed during training. \cite{zhao2020learning} learns to discover geometrically consistent 3D keypoints from pairs of images of object with rigid body transformation information. \citet{tang2019self} learns depth-aware keypoints through appearance and geometric matching from videos for autonomous driving. 

\vspace{-0.05in}
\paragraph{Keypoint Discovery for Control}
\citet{florence2018dense} propose to learn an object-centric dense correspondence model through contrastive learning. The process involves a dataset collected without supervision, by scanning target object from a variety of angles using a robotic arm and doing 3D reconstruction using the data. \cite{manuelli2020keypoints} samples points from this dense correspondence model as descriptor of an object and applies model predictive control on such keypoint based descriptor. \cite{manuelli2019kpam} proposes to use semantic 3D keypoints as the object representation for category-level pick and place tasks instead of transformation from a fixed object template. This method, however, requires manually labeled keypoint dataset. \citet{gao2019kpam} improves its result by combining these keypoints with dense geometry of object. In contrast, we learn 3D keypoints directly from policy learning loss in an unsupervised and end-to-end manner.

\section{Conclusion}
In this work, we presented a framework to learn useful 3D keypoints without supervision for continuous control. Our key insight is to leverage multi-view consistency with a world coordinate transform in the bottleneck layer for learning reliable keypoints. We jointly train unsupervised 3D keypoint learning in conjunction with reinforcement learning to achieve significant sample efficiency improvement in a variety of 3D control environments. The 3D keypoints learned by our algorithm are consistent across space and time. The keypoints offer several benefits as they make the control policy less dependent on the visual input, and hence, provide a great means for transfer learning even if the visual distribution changes a lot, for instance, transfer from simulation to real. We hope our method serves as a bridge between pixel domain and 3D control tasks.

\section*{Acknowledgments}
We thank Zackory Erickson, Alexander Clegg, and Charlie Kemp for fruitful discussions. This work was supported by NSF IIS-2024594 and NSF IIS-2024675.

\bibliographystyle{icml2021}
\bibliography{main}
\appendix
\section{Additional Implementation Details}
\paragraph{Keypoint Resampling} We find that the decoder can potentially cheat by hiding the input information in the relative locations keypoint to each other.
To handle this issue, we do not use the exact locations produced as output by the keypoint encoder but re-sample them from the empirical distribution.
% At inference time, $[\mathbb{E}[u_{n}^{k}],\mathbb{E}[v_{n}^{k}],\mathbb{E}[d_{n}^{k}]]^\top$ is the coordinate of the $k$-th keypoint in the $n$-th camera frame that is used downstream.
We find that during training, injecting noise by re-sampling around $[\mathbb{E}[u_{n}^{k}],\mathbb{E}[v_{n}^{k}],\mathbb{E}[d_{n}^{k}]]^\top$ (corresponding to $k$-th keypoint in the $n$-th camera frame) produces sharper keypoint heatmaps that avoids potential multimodality issues. Such noise also encourages keypoints to correspond to actual visual features rather be simply used as bits to pass information to decoder. We can calculate $\sigma_{n}^{k}$, the spatial standard deviation of $[u_{n}^{k},v_{n}^{k}]^\top$, from the heatmap. 
\begin{eqnarray*}
\sigma_{n}^{k}
&=&\sqrt{\sum_{u=1}^{S}\sum_{v=1}^{S}||\begin{bmatrix} u/S\\ v/S\end{bmatrix}-\begin{bmatrix} \mathbb{E}[u_{n}^{k}]\\ \mathbb{E}[v_{n}^{k}]\end{bmatrix}||_{2}^2\cdot H_{n}^{k}(u,v)}
\end{eqnarray*}
At training time, we resample $[\hat{u}_{n}^{k}, \hat{v}_{n}^{k}, \hat{d}_{n}^{k}]^\top=[\mathbb{E}[u_{n}^{k}]+a\sigma_{n}^{k},\mathbb{E}[v_{n}^{k}]+b\sigma_{n}^{k},\mathbb{E}[d_{n}^{k}]]^\top$ where $a,b \sim \mathcal{N}(0,\nu)$ for some positive noise hyper parameter $\nu$. This resampled version of predicted keypoint coordinate encourages heatmap to have small spatial variance to minimize noise while also making decoder more robust to noisy keypoint predictions. At inference time, we set $\nu=0$ to disable noise and compute $[\hat{u}_{n}^{k}, \hat{v}_{n}^{k}, \hat{d}_{n}^{k}]^\top$=$[\mathbb{E}[u_{n}^{k}],\mathbb{E}[v_{n}^{k}],\mathbb{E}[d_{n}^{k}]]^\top$.

\paragraph{First Frame Features}
When reconstructing observed images from distillated keypoints, the decoder needs to recover the relatively static background pixels. This can be achieved by decoder itself. However, doing so wastes the compute and parameters of the decoder. We thus use the same technique in the 2D Keypoint baseline \cite{minderer2019unsupervised} in experiment section, to alleviate this problem by concatenating first frame features to the Gaussian maps before decoding. Such features are extracted by passing a canonical frame, such as the first frame of environment or the photo of an empty scene, to a small convolutional encoder. This small encoder effectively captures the static pixels for reconstruction so the decoder can focus on keypoints better. 

\begin{table}
\centering
\begin{tabular}{lll}
\toprule
\textbf{Hyperparameter} & Metaworld & Ant  \\
\midrule
PPO batch size (metaworld) & 6400 & 16000 \\
Rollout buffer size & 100000 & 100000\\
\# Epochs per update & 8 & 10\\
gamma & 0.99 & 0.99\\
gae lambda & 0.95 & 0.95\\
clip range ($\epsilon$) & 0.2& 0.2\\
entropy coefficient & 0.0& 0.0\\
value function coefficient & 0.5& 0.5\\
gradient clip &0.5 &0.5\\
target KL & 0.12& 0.12\\
policy learning rate & $3e-4$& $3e-4$ \\
observation buffer size & 100000 & 100000\\
unsupervised learning rate & $3e-4$ & $3e-4$\\
\# keypoints & 32& 16\\ 
\# cameras (N) & 3& 3\\ 
\# unsupvised learning steps (p) & 400& 400\\ 
noise $\nu$ when no augmentation & $5e-4$ & $5e-4$ \\
noise $\nu$ when augmented& 0.0 & 0.0\\
Autoencoding weight ($\lambda_{ae}$)& 5.0& 5.0\\ 
Multiview weight ($\lambda_{multi}$)& 0.05& 0.05\\ 
Seperation weight ($\lambda_{sep}$)& 0.0025& 0.0025\\ 
\bottomrule
\end{tabular}
\caption{The values of all hyperparameters for our algorithm. The code is available at~\url{https://buoyancy99.github.io/unsup-3d-keypoints/}.}
\label{tab:hp}
\end{table}

\paragraph{Additional states}
In partially observable environments, we also concatenate some unobservable robot states with learned keypoints before feeding into the fully connected policy layers. These states are those easily accessible on real robot but can hardly be assessed from our third person view camera setup. For metaworld, we use an indicator of whether the gripper is opening or closing because the gripper is hardly visible when arm is far from the third-person-view cameras. For pybullet-ant, such state is the x coordinate of the robot center, which cannot be estimated visually since the camera is always moving with robot and the ground is white. In scarf environment, such state is pr2 robot's arm joint angles since they can hardly be estimated under low resolution even by humans. They are also critical states for joint space control with obstacle avoidance.

\paragraph{Implementation details}
We provide all hyper parameters for PPO training as well as that for unsupervised learning in \cref{tab:hp}. More details can be found in the code we provided. 

\begin{figure}[t]
\centering
\includegraphics[width=\linewidth]{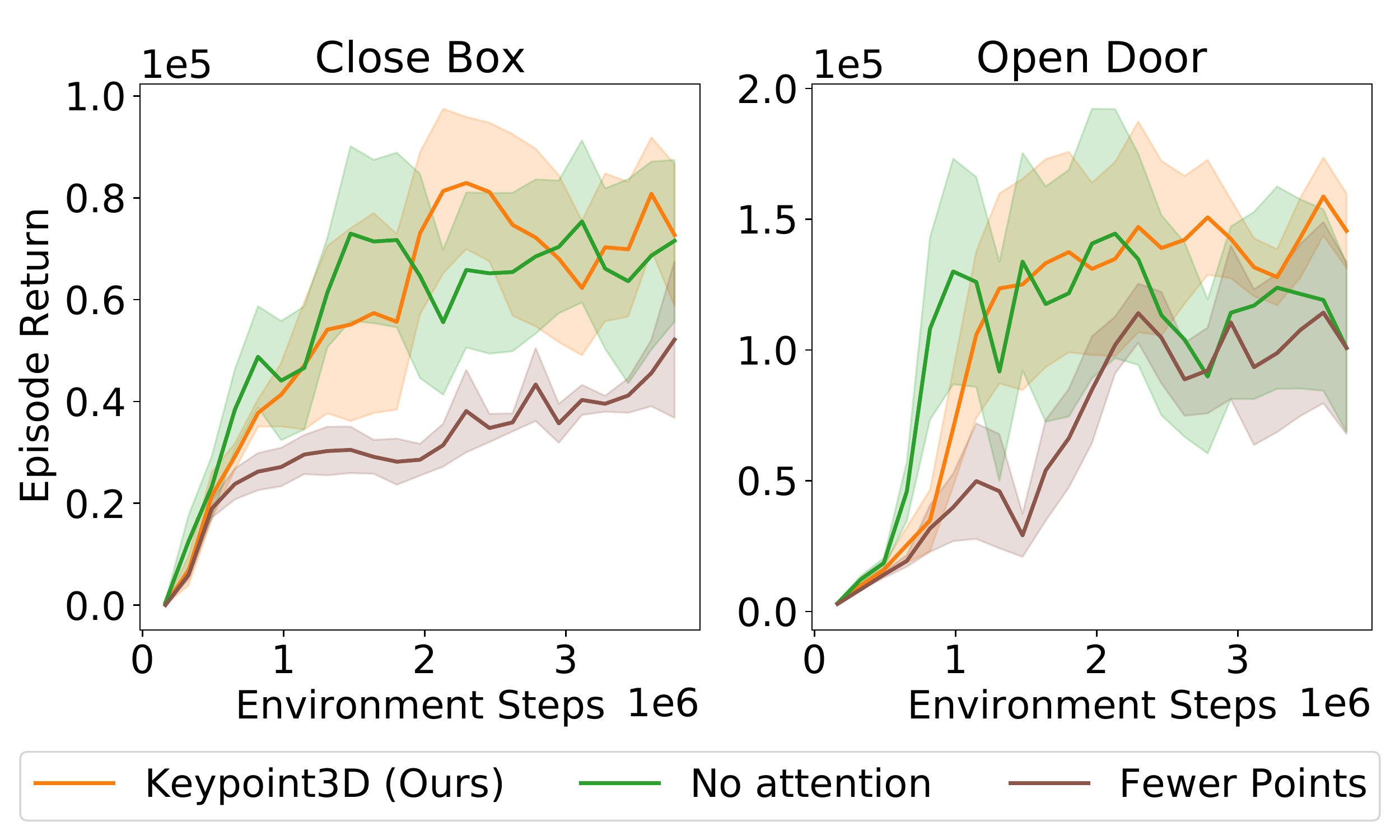}
\vspace{-0.25in}
\caption{Additional ablations of our Keypoint3D approach with respect to different design choices. Variations include: removing attention, using fewer keypoints.}
\label{fig:supp_ablate_curve}
\end{figure}

\section{Additional Ablations}
In addition to the ablations provided in the main paper, we present few more ablative studies in this section for better understanding of our method.

\paragraph{Effects of attention}
\cref{fig:supp_ablate_curve} shows removing the attention mechanism has some minor impact on the reinforcement learning performance. To understand this, we notice the role of attention is to ignore irrelevant keypoints. To achieve this, we can either use attention or let the decoder learn to figure out by itself. Thus the decoder can still achieve the similar effect when attention is removed. However, we still decide to keep the attention module since we found attention mechanism to be very important for visualization of keypoints, in which we use a threshold on attention to filter out unimportant keypoints. 

\paragraph{Effects of Effects of number of keypoints}
We already used much fewer keypoints compared to prior unsupervised learning methods~\cite{minderer2019unsupervised}. However, to further investigate the effect of this problem on keypiont-based methods by drastically decrease the number of keypoint used. We use $\frac{3}{16}$ of the number of points used in our original method and notice drop in performance as indicated in \cref{fig:supp_ablate_curve}.

\section{Additional Visualizations}
\begin{figure*}[t]
\centering
\begin{subfigure}[b]{\linewidth}
\includegraphics[width=\linewidth,height=0.3\linewidth]{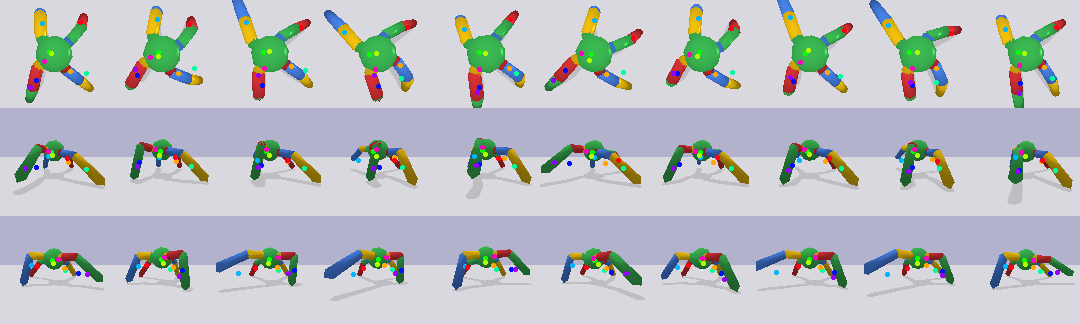}
\caption{In pybullet ant, the learned keypoints track limbs of the locomotion robot}
\label{fig:antsupp}
\end{subfigure}
\begin{subfigure}[b]{\linewidth}
\includegraphics[width=\linewidth,height=0.3\linewidth]{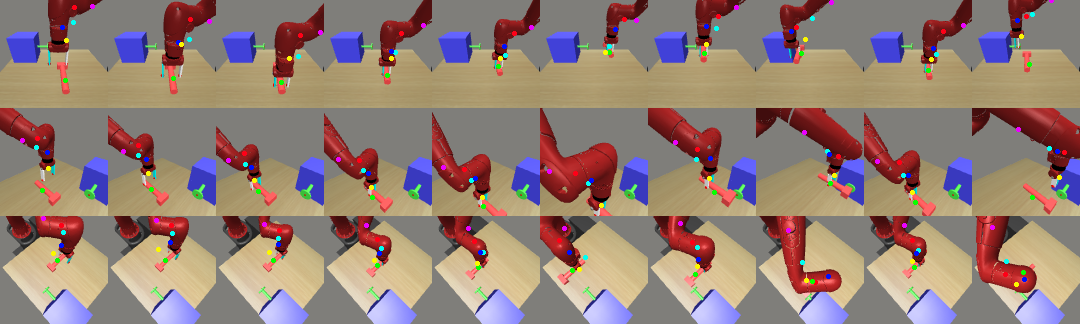}
\caption{In hammer manipulation, the learned keypoints track joints of arms as well as the hammer}
\label{fig:hammersupp}
\end{subfigure}
\begin{subfigure}[b]{\linewidth}
\includegraphics[width=\linewidth,height=0.3\linewidth]{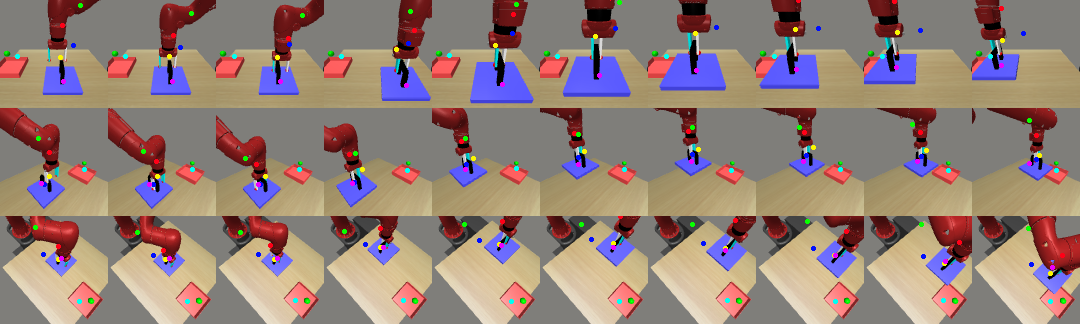}
\caption{In box closing, the learned keypoints track the cover, the end effector and the box}
\label{fig:bcsupp}
\end{subfigure}
\caption{Our unsupervised learning based Keypoint3D method learns high-quality 3D keypoints across the benchmarked manipulation tasks. We provide video visualizations of the 3d keypoints on the website~\url{https://buoyancy99.github.io/unsup-3d-keypoints/}. Please refer to videos for better understanding of results.}
\label{fig:vissupp}
\end{figure*}

In \cref{fig:vissupp}, we provide additional visualizations on several environments. In the pybullet ant environment as shown in \cref{fig:antsupp}, we can observe that the red point tracks the movement of upper red leg very consistently in all poses. In \cref{fig:hammersupp}, a the green point tracks the handle of the hammer, whether it's on table top or in the gripper. Other points follow different joints of the robot arm respectively. When the hammer drops onto the table in the right most column, the point movement also reflects this change by moving itself away from those points corresponding to the arm. In \cref{fig:bcsupp}, the purpose point moves with the cover of the box while the red point and yellow point follows different segments the end effector. Overall, our unsupervised learning method learns high-quality meaningful 3D keypoints for these 3D tasks.

\end{document}